# A Systematic Review of Machine Learning Approaches for Detecting Deceptive Activities on Social Media: Methods, Challenges, and Biases

Yunchong Liu[1], Xiaorui Shen[2], Yeyubei Zhang[1], Zhongyan Wang[3], Yexin Tian[4], Jianglai Dai[5], and Yuchen Cao[2]

[1]School of Engineering and Applied Science, University of Pennsylvania

[2]Khoury college of computer science, Northeastern University

[3]Center for Data Science, New York University

[4]Georgia Institute of Technology, College of Computing

[5]Department of EECS, University of California, Berkeley

**Abstract**

Social media platforms like Twitter, Facebook, and Instagram have facilitated the spread of misinformation, necessitating automated detection systems. This systematic review evaluates 36 studies that apply machine learning (ML) and deep learning (DL) models to detect fake news, spam, and fake accounts on social media. Using the Prediction model Risk Of Bias ASsessment Tool (PROBAST), the review identified key biases across the ML lifecycle: selection bias due to non-representative sampling, inadequate handling of class imbalance, insufficient linguistic preprocessing (e.g., negations), and inconsistent hyperparameter tuning. Although models such as Support Vector Machines (SVM), Random Forests, and Long Short-Term Memory (LSTM) networks showed strong potential, over-reliance on accuracy as an evaluation metric in imbalanced data settings was a common flaw. The review highlights the need for improved data preprocessing (e.g., resampling techniques), consistent hyperparameter tuning, and the use of appropriate metrics like precision, recall, F1 score, and AUROC. Addressing these limitations can lead to more reliable and generalizable ML/DL models for detecting deceptive content, ultimately contributing to the reduction of misinformation on social media.

*Keywords:* fake news, deceptive activities, social media, bias evaluation with PROBAST, systematic review, machine learning, deep learning

**Introduction**

Over the past two decades, social media platforms such as Twitter, Facebook, Instagram, and Reddit have dramatically reshaped how people communicate, share information, and engage with global events (Smith & Anderson, 2018). While these platforms provide unparalleled opportunities for connection and information dissemination, they have also become fertile ground for the spread of misinformation, fake news, spam, and fraudulent activities (Vosoughi, Roy, & Aral, 2018). The rise in deceptive content has far-reaching societal consequences For example, during the 2016 U.S. presidential election, widespread dissemination of false political content is believed to have influenced voter behavior, with research indicating that this type of content played a significant role in shaping public opinion (Allcott & Gentzkow, 2017). Similarly, the COVID-19 pandemic saw social media become a hotbed for health-related misinformation. False information about vaccines and the virus spread rapidly, contributing to public confusion, increased vaccine hesitancy, and the undermining of global public health efforts. (Zarocostas, 2020). In addition to these, misinformation has also played a role in inciting social unrest. For instance, during the Russia–Ukraine war, false narratives spread through Turkish social media platforms contributed to heightened tensions and public panic. These fake posts fueled divisive narratives, exacerbating the conflict's impact on local and international communities (Uluşan & Özejder, 2024).

In addition to fake news, fake accounts, including bots, spam accounts, and identity theft profiles, have proliferated across social media platforms. These accounts are frequently used to spread misinformation or carry out fraudulent activities, exacerbating the already significant problem of deceptive content online. Bots, in particular, can artificially amplify the reach of fake news by inflating engagement metrics such as likes, shares, and retweets, creating the illusion of widespread belief in false information. This issue became especially pronounced during the COVID-19 pandemic, where bots played a crucial role in spreading health misinformation, further complicating public understanding and responses to the crisis (Himelein-Wachowiak et al., 2021).

One of the key challenges in combating misinformation is the sheer volume of data generated on social media. Every day, millions of posts, tweets, images, and videos are shared, making manual monitoring and verification impossible (Shu et al., 2017). An even bigger challenge lies in the speed and scale at which misinformation spreads. Social media algorithms, designed to prioritize user engagement, frequently promote sensational or controversial content, regardless of its accuracy (Vosoughi et al., 2018, Chen et al., 2025). These algorithms are structured to maximize time spent on platforms, making the spread of false information an unintended but inherent consequence of their design. As a result, false information tends to spread faster than the truth, especially in contexts like political elections or public health emergencies. A study by Vosoughi, Roy, and Aral (2018) found that false news stories on Twitter were 70% more likely to be retweeted than true stories. This tendency is not accidental but is embedded in the business model of platforms like Facebook, which profit from increased engagement with sensational content (Lauer, 2021).

Traditional moderation methods, which often rely on human intervention, are insufficient to address the scale and speed at which deceptive content spreads. As such, automated, efficient, and accurate methods to detect and mitigate the impact of deceptive content are essential to safeguard information integrity online (Lazer et al., 2018).

In response to these challenges, machine learning (ML) and deep learning (DL) techniques have emerged as powerful tools for detecting deceptive content on social media platforms (Zhou & Zafarani, 2020). ML algorithms, such as Support Vector Machines (SVM), Random Forests, and Logistic Regression, have been employed to classify content based on features extracted from textual, visual, and network data (Conroy, Rubin, & Chen, 2015). These models are highly effective at identifying patterns in the data that distinguish between legitimate and deceptive content, but they often rely on handcrafted features, which can limit their adaptability to new types of misinformation.

To address this limitation, deep learning (DL) models have significantly advanced detection capabilities, including Convolutional Neural Networks (CNNs) and Recurrent Neural Networks (RNNs), can automatically learn complex patterns and representations from large amounts of unstructured data, such as text and images (Ruchansky, Seo, & Liu, 2017). These models eliminate the need for manual feature extraction, making them more versatile in adapting to evolving forms of misinformation.

In particular, natural Language Processing (NLP) models play a crucial role in analyzing textual content. By detecting subtle linguistic cues and anomalies that indicate fake news or spam, NLP models are particularly useful for processing large volumes of data in real time (Zhang & Ghorbani, 2020). These models can dynamically adapt to shifting patterns of deceptive behavior, which is essential for keeping pace with the rapid spread of misinformation.

The integration of ML and DL techniques provides a scalable solution to identify and filter out malicious content, thereby safeguarding the integrity of information on social media platforms. These automated approaches provide the efficiency and accuracy needed to combat misinformation on a large scale, where manual methods fall short (Shu et al., 2017).

Challenges of Using Machine Learning/Deep Learning Models in Detecting Deceptive Content

Despite the significant advancements in the application of ML and DL models for detecting misinformation and deceptive activities on social media, several challenges hinder the full effectiveness and scalability of these approaches. These challenges arise from the complex nature of social media data, technical limitations of the models, and broader concerns about bias and ethical use.

One of the fundamental challenges arises from the complexity of social media data itself. Social media content is highly unstructured and multimodal, encompassing text, images, videos, and metadata that vary across users and platforms. Posts are not uniform; they often include slang, abbreviations, emojis, and varying linguistic styles, making traditional text-processing techniques less effective (Kaplan & Haenlein, 2010; Li & Li, 2014; Dan et al., 2024). Additionally, the context of posts plays a critical role in determining whether it is deceptive or legitimate, but algorithms typically struggle to interpret such nuances. For instance, a post that is sarcastic, humorously exaggerated, or part of an ongoing conversation can be mistakenly flagged as misinformation if the model does not accurately capture the context (Gupta & Kumaraguru, 2012). Moreover, the dynamic and rapidly evolving nature of social media further adds to the complexity, as models must adapt to new forms of deception and manipulation techniques (Ferrara et al., 2016). One such example is the proliferation of social bots, a type of fake account designed to simulate human behavior. Social bots are often used to distort online discussions and amplify misinformation. During the 2016 U.S. Presidential Election, social bots were deployed to artificially inflate political discourse, giving the illusion of widespread support or opposition and ultimately influencing public perception (Bessi & Ferrara, 2016, Gao et al., 2024). The increasing sophistication of such fake accounts presents a significant challenge for detection models, which must continuously evolve to keep pace with new forms of automated deception.

Another major issue is the class imbalance inherent in the detection of misinformation and fake content. One of the most significant technical challenges in the application of ML and DL models to misinformation detection is the class imbalance problem. Legitimate content vastly outnumbers fake or deceptive content, leading to datasets where the minority class (e.g., fake news, spam, or bot accounts) is severely underrepresented (Chawla, Japkowicz, & Kotcz, 2004). This imbalance skews the model's learning process, causing it to prioritize accuracy on the majority class—legitimate content—while failing to correctly identify the minority class of interest, such as fake accounts or misinformation (He & Garcia, 2009). As a result, the models may exhibit high overall accuracy but poor sensitivity to the very content they are designed to detect. Addressing class imbalance is essential to improve the sensitivity of models to deceptive activities without compromising their specificity (Sun, Wong, & Kamel, 2009).

Furthermore, generalization is a persistent problem. ML and DL models often struggle with generalization—the ability to perform well on data that differs from the training dataset. Many models are trained on specific datasets from certain social media platforms or under controlled conditions, which limits their ability to generalize across platforms or adapt to new contexts. This issue is particularly evident in deep learning models, which, despite their capacity to learn complex patterns, can be overfit to the training data and perform poorly when applied to new or unseen data (Bay & Yearick, 2024, Li et al., 2024). For instance, a model trained on Twitter data may not perform as effectively on Reddit or Facebook, where user behavior and content structures differ significantly. Additionally, recent studies have highlighted that deep learning models, especially large language models, are vulnerable to adversarial attacks—intentionally designed inputs aimed at misleading the model's predictions—further underscoring the need for robust evaluation methods to ensure reliability in real-world misinformation detection scenarios (Tao et al., 2024).

Objective of the Review

This systematic review aims to critically evaluate the application of machine learning and deep learning techniques in detecting misinformation, spam, fake profiles, and other deceptive activities on social media platforms. By analyzing studies published between January 2010 and July 2024, the review seeks to:

- Assess the effectiveness of various ML and DL models employed in this domain.
- Identify common challenges and limitations faced by these models, including issues related to data complexity, class imbalance, generalization, and other methodological biases.
- Propose future directions for research to enhance model performance, generalizability, and ethical considerations.

Through this comprehensive evaluation, the review intends to contribute to the development of more robust, accurate, and ethically responsible detection methods. By addressing existing gaps and highlighting best practices, it aims to support efforts in mitigating the spread of misinformation and enhancing the reliability of information shared on social media platforms.

The subsequent sections will outline the methodologies employed in the reviewed studies, present the results of this analysis, and discuss the implications of these findings for future research and practice. We will also explore the potential for improving model performance and reliability in real-world applications of ML and DL to combat misinformation on social media.

**Method**

Search Strategy

This review focused on identifying studies that utilized machine learning (ML) and deep learning (DL) techniques to detect misinformation, spam, fake profiles, and other deceptive activities on social media platforms such as Twitter, Facebook, Instagram, and Reddit. These platforms were chosen due to their prevalence in social discourse and their known issues with the spread of deceptive content, making them key areas for research on misinformation detection. The search terms included combinations of "machine learning", "deep learning", "Twitter", "Facebook", "Instagram", "Reddit", "fraud", "fake account", and "fake content." The focus on these terms aimed to capture a broad range of deceptive activities and the ML/DL techniques used for their detection.

The comprehensive search covered multiple academic databases, including five databases, PubMed, Google Scholar, IEEE Xplore, ResearchGate, and ScienceDirect, chosen for their extensive coverage of peer-reviewed literature relevant to both technical and medical fields. These databases were selected to ensure that studies from a wide array of disciplines, including computer science, data science, and social media analysis, were included. This extensive search was designed to capture a wide range of studies relevant to research objectives. The search process was carried out from June to July 2024.

The search terms were carefully selected to cover the key aspects of the studies under review, including the social media platforms under investigation, the types of deceptive activities being detected, and the machine learning techniques used. The following terms were applied:

1. Social Media Platforms: "Twitter", "Facebook", "Instagram", and "Reddit".
2. Deceptive Activities: "fake news", "spam", "fraud", "fake accounts", "faked content", and "risk".

3. Machine Learning Techniques: "machine learning", "deep learning", "neural networks".

The search query used for this review was: (("Twitter" OR "Facebook" OR "Instagram" OR "Reddit") AND ("fake news" OR "spam" OR "fraud" OR "fake accounts" OR "fake content" OR "risk") AND ("machine learning" OR "deep learning" OR "neural networks")).

Inclusion/Exclusion Criteria and Study Selection Process

Studies were included in this review if they met the following criteria:

- Publication Date: Studies published from January 2010 to July 2024 were selected to capture the most recent developments in ML and DL applications for detecting deceptive activities.
- Language: Only studies published in English were considered, ensuring accessibility and consistent analysis across the literature.
- Research Focus: Studies that applied ML and DL techniques specifically to detect various forms of deceptive behavior (e.g., misinformation, spam, fake accounts) on social media platforms were included.
- Social Media Data: Studies that analyzed posts from popular social media platforms like Twitter, Reddit, Instagram, and Facebook, were eligible, ensuring that the focus remained on widely used platforms with prevalent deceptive activities.
- Event-Related Data: Studies analyzing data related to specific events (e.g., crises, crimes, or topics to understand the context of deceptive activities) were included.
- Relevant Datasets: The inclusion criteria required the use of datasets relevant to the detection of deceptive content, such as those focusing on spam detection, social media misbehavior, or fake account detection.
- Study Design: Primary research articles employing observational or experimental methodologies were included, as well as studies utilizing ML/DL algorithms for detection purposes.

Studies were excluded if they fell into the following categories:

- Publication Type: Non-peer-reviewed articles, review articles, and conference abstracts were excluded to ensure the focus remained on original, peer-reviewed research. Irrelevant/Duplicates: Studies with non-relevant focus or those identified as duplicates were excluded. Studies based on private social media accounts were also excluded due to data privacy concerns.

- Insufficient Information: Studies lacking sufficient information for classification or those without detailed methodology (e.g., missing data sources, algorithms) were excluded.

- Language Restrictions: Studies Published in languages other than English were excluded.

- Incomplete Data: Studies with incomplete or missing data that prevents classification or analyses, were excluded.

- Scope: Studies that did not focus on social media or failed to use ML/DL techniques for detection of deceptive activities were excluded.

The study selection process involved several structured steps to ensure the relevance and quality of included studies:

1. Initial Identification: Duplicates were removed, and preliminary screening was conducted based on titles and abstracts to determine relevance.

2. Title and Abstract Screening: Two reviewers independently assessed the studies to ensure they met the inclusion and exclusion criteria. Any differences were addressed and resolved through discussion to maintain consistency in the selection process.

3. Full-Text Screening: A thorough review of the full texts was performed, with any discrepancies addressed and resolved through further discussion.

4. Data Extraction and Final Inclusion: Relevant information was extracted from the selected studies. The final list of included studies was established, ensuring they meet all criteria and are aligned with the research objectives.

5. Documentation: The entire selection process was documented, including reasons for exclusion at each stage and the final list of studies, ensuring transparency and reproducibility of the research.

Data Extraction and Analysis

The data extraction process was carried out by systematically gathering comprehensive information from each included study through a standardized template. This template captured crucial details such as the authors, study titles, journals of publication, and publication dates, as well as information on study design, settings, sample sizes, and inclusion/exclusion criteria. In addition to this, the template tracked the ML and DL models utilized, the social media platforms examined (e.g., Twitter, Facebook, Instagram), and the primary and secondary outcomes observed in each study. Performance metrics, including accuracy, precision, recall, F1 score, and Area Under the Receiver Operating Characteristic Curve (AUROC), were extracted when reported. Moreover, potential biases, study limitations, and funding information were documented to ensure a thorough understanding of the study's reliability and context.

---

Insert Table 1 about here

---

The review examined the types of data and content analyzed in the studies, such as distinctions between public and private posts, language-specific data, and the relevance of content to specific topics. This approach provided a nuanced understanding of how different data sources and contexts impact the performance and applicability of ML/DL models used for detecting misinformation and spam across various social media platforms.

To ensure a thorough evaluation of bias, an established tool, the Prediction model Risk Of Bias ASsessment Tool (PROBAST) (Wolff et al., 2019), was employed. This tool helped assess the credibility and reliability of findings by identifying where models might be overfitting or underperforming due to dataset limitations.

A critical part of the evaluation was assessing key performance metrics—such as accuracy, precision, recall, F1 scores, and AUROC—particularly in relation to challenges like imbalanced data. These metrics were essential for understanding how well the models performed in detecting deceptive content, given the disproportionate presence of legitimate versus fake content on social media. This analysis highlighted how biases in model training, such as favoring majority classes, could undermine model effectiveness in identifying minority-class events like misinformation or spam.

The review also identified other research gaps, such as how well these models generalize to diverse real-world social media environments. By integrating performance metrics and bias assessments, this study aimed to provide a comprehensive overview of current research, offering insights into areas for improving model accuracy, handling imbalanced data, and ensuring wider applicability in detecting misinformation and other deceptive activities on social media platforms.

## Results

### Study Selection

A comprehensive literature search was conducted across multiple databases to identify relevant studies on machine learning models used for fake news detection on social media platforms. The databases searched included PubMed, IEEE Xplore, ScienceDirect, and Google Scholar. The search terms used were combinations of keywords such as "machine learning," "Twitter," "fake," "fraud," "spam," "Reddit," "Facebook," "Instagram," and "deep learning." The diagram in Figure 1 illustrates the study selection process based on the search results.

---

Insert Figure 1 about here

---

The comprehensive literature search resulted in 903 studies across four major databases: IEEE Xplore contributed the largest portion with 361 studies, followed by Google Scholar with 271 studies, PubMed with 181 studies, and ScienceDirect with 90 studies. After removing 98 duplicate studies, a total of 805 unique titles and abstracts were subjected to initial screening. During this phase, 690 studies were excluded based on several exclusion criteria: 241 studies were deemed irrelevant to the research topic, 172 used inappropriate data sources, 138 applied unsuitable techniques, 104 focused on unrelated disorders, and 34 were excluded due to publication type.

The remaining 115 studies underwent full-text screening, which resulted in the exclusion of 79 additional publications. Reasons for exclusion at this stage included irrelevance (39 studies), issues with data sources (3 studies), inappropriate techniques (3 studies), unrelated disorders (20 studies), unsuitable publication type (8 studies), language limitations (1 study), and the unavailability of data (5 studies). Ultimately, 36 studies met the inclusion criteria and were included in the narrative synthesis for the systematic review, as shown in the diagram in Figure 1.

Characteristics of Included Studies

The studies reviewed in this systematic analysis focused on detecting fake news and deceptive activities across various social media platforms. Among the 36 studies included, Twitter was the most frequently analyzed platform, appearing in 19 studies. Instagram was the subject of 9 studies, and both Facebook and Reddit were analyzed in 3 studies each. Additionally, 1 study focused on Weibo, a Chinese platform like Twitter, while 2 studies explored data from multiple platforms, including Twitter, Facebook, YouTube, and email. This reflects the diverse environments where misinformation and fake profiles are commonly found.

The types of deceptive content analyzed in these studies were equally varied. The most prevalent focus was on fake profiles and fake accounts detection, which was explored in 15 studies. Another significant area of interest was the general detection of fake news, addressed in 12 studies. Spam and scam detection was examined in 5 studies, while phishing detection was the focus of 2 studies. There were also 2 studies that addressed health-related misinformation, such as detecting signs of suicidal ideation, and 1 study that focused specifically on misinformation spread during crisis events like natural disasters or terrorist attacks.

A variety of machine learning and deep learning models were employed across the studies. Supervised machine learning models were particularly popular, with Random Forest being the most widely used, appearing in 17 studies. Support Vector Machines (SVM) were applied in 16 studies, while Naive Bayes and Logistic Regression were used in 12 and 14 studies, respectively. Other models, such as Decision Trees, K-Nearest Neighbors (KNN), Gradient Boosting, and Stochastic Gradient Descent (SGD), were also featured, though less frequently.

Deep learning models played a prominent role in several studies, with Artificial Neural Networks (ANN) and Multi-layer Perceptron (MLP) employed in 8 studies. Convolutional Neural Networks (CNN) and Recurrent Neural Networks (RNN) were each used in 2 studies, and Long Short-Term Memory (LSTM) models were also applied in 2 studies. Additionally, autoencoders and hybrid models, which combine techniques such as CNN with RNN or other machine learning approaches, were explored in 4 studies. Transformer-based models and deep stacked autoencoders, representing the latest advancements in deep learning, were used in a smaller number of studies.

In summary, the studies demonstrated a strong reliance on both traditional supervised machine learning models like Random Forest and SVM, as well as more advanced deep learning models such as ANN and LSTM. These models were particularly effective in detecting fake news and fake profiles, with high accuracy rates reported across multiple studies. This comprehensive review highlights both the diversity of approaches and the growing complexity of model usage in the field of misinformation detection on social media platforms.

Methodological Quality and Risk of Bias

We employed the Prediction model Risk Of Bias ASsessment Tool (PROBAST) (Wolff et al., 2019) to systematically evaluate potential biases across four domains: sample selection and representativeness, data preprocessing, model development, and model evaluation. Each domain was examined through a series of targeted questions, as shown in Table 2. These questions helped identify biases that could affect the reliability and generalizability of the machine learning models. Key concerns included the representativeness of samples, the handling of negative words, hyperparameter tuning, and the appropriateness of evaluation metrics, especially in class-imbalanced settings. This structured approach allowed for a clear identification of the strengths and limitations of the studies, ensuring a rigorous assessment of methodological quality.

---

Insert Table 2 about here

---

Sample Selection and Representativeness (Q1 & Q2):

The reviewed studies employed various sampling methods across multiple social media platforms, including Twitter, Instagram, Facebook, and Reddit, aiming to investigate phenomena such as misinformation, fake accounts, or spam detection. However, only a small fraction of the studies aimed to provide representative samples of the broader social media user base, with most research focusing on particular events, user groups, or trends. This limits the generalizability of their findings, as they fail to capture the broader behaviors of social media users. For instance, some studies (e.g., Study #1) specifically targeted tweets during crisis situations. These approaches do not reflect the larger population of social media users. Conversely, while others (e.g., Study #4) employed random sampling methods that were somewhat more representative but remained confined to Twitter. Studies like these do not reflect the broader social media landscape. A detailed summary of the sampling approaches used in the studies is provided in Table 3.

Among the 36 studies reviewed, 20 studies (55%) focused on specific events, user groups, or phenomena without attempting to represent the broader social media population. For example, Study #1 focused solely on Twitter, collecting 15,952 tweets related to six specific misinformation events during crisis periods, using data gathered within specific time frames to monitor misinformation spread. Another study is Study #2, which concentrated on Instagram by web scraping data from the platform, resulting in a dataset of 970 bot accounts, 959 real accounts, and 870 fake accounts. Similarly, Study #3 utilized Twitter data collected from over 54 million user accounts but narrowed the focus to tweets associated with three trending topics, specifically targeting and manually labeling tweets as spammers based on keyword presence.

Additionally, 10 studies (28%) employed sampling methods that were more general but remained constrained by platform specificity, typically focusing on a single platform such as Twitter or Instagram. For instance, Study #4 used a dataset of tweets where phishing content was manually identified by security experts, while non-phishing tweets were randomly sampled from the Twitter stream.

The remaining 6 studies (17%) attempted more comprehensive or representative sampling across multiple social media platforms or utilized datasets aimed at being more inclusive, but they too faced limitations related to platform-specific constraints like language and geography. Study #22, for instance, utilized over 15,000 news contents from Facebook, collecting through a custom-built crawler and the Facebook API to gather public user information and posts and analyze both fake and real news. Study #11 took a more inclusive approach, using data from both Reddit and Cable News Network (CNN) to improve fake news detection.

---

Insert Table 3 about here

---

Non-representative sampling introduces selection bias, which can undermine the generalizability of the study's findings. Studies that focus exclusively on certain social media platforms or specific user groups may overlook significant variations in user behavior, interactions, content, and platform-specific dynamics across the broader social media landscape. For example, a study focusing solely on tweets from specific crisis events or narrowly targets categories like fake accounts may fail to capture the full spectrum of behaviors present on other platforms like Facebook or Reddit. This limitation compromises the validity of the conclusions, especially when attempting to apply the findings to a general population. Fortunately, nearly 95% of the reviewed papers acknowledge these limitations, demonstrating an awareness of the inherent challenges posed by non-representative sampling, particularly in the field of ML/DL applications for social media analysis. In the broader machine learning literature, researchers such as Goodfellow, Bengio, and Courville (2016) underscore the importance of representative sampling to build models that generalize effectively and avoid the risk of overfitting to specific data subsets, thereby enhancing the reliability and applicability of findings across diverse social media contexts.

This analysis reveals a significant challenge of achieving representative sampling in ML/DL studies applied to social media platforms. The prevalent reliance on non-representative samples introduces selection bias, limiting the validity and generalizability of the findings. Even studies that incorporate multiple platforms struggle with achieving true representativeness due to constraints tied to linguistic, geographical, and demographic diversity. This challenge aligns with findings from Shen et al. (2025), who demonstrated that limited-resource language contexts pose difficulties for generalization, further complicating efforts in building broadly applicable misinformation detection models. Most authors acknowledge these limitations, recognizing that fully representative sampling is difficult to achieve given the sheer diversity of the social media landscape. Moving forward, a critical future direction is to develop more inclusive sampling strategies that incorporate multiple platforms and strive to account for factors such as language, geography, and user demographic. By doing so, future research can more effectively capture the complexities of the social media ecosystem, enhancing the robustness of ML/DL models and improving the applicability of their conclusions across diverse social media contexts.

Data Preprocessing with Focus of Negative Words Handling (Q3)

Effective data preprocessing is pivotal in machine learning workflows, particularly in tasks like fake news detection, where linguistic nuances—such as negations—can have significant influence on model performance. This section evaluates how the reviewed studies addressed data preprocessing tasks, focusing specifically on the treatment of negative words, which is crucial in accurately interpreting textual content.

In the reviewed studies, preprocessing techniques to prepare textual data for machine learning models followed consistent patterns. One common step involved normalizing text, which included converting all characters to lowercase and removing punctuation, URLs, and special symbols. Tokenization, the process of breaking text into individual units or tokens, was universally employed across studies. Some studies also applied stemming and lemmatization techniques to reduce words to their base or root forms, ensuring uniformity across different grammatical variants. Another frequently used step was the removal of stop words, eliminating commonly used words that add little value to modeling. To convert textual data into numerical representations for model input, feature extraction methods such as Term Frequency-Inverse Document Frequency (TF-IDF), Bag of Words (BoW), and word embeddings were widely adopted (Singh & Singh, 2022).

Of the 36 reviewed studies, 30 applied traditional machine learning methods using token-based features like n-grams and TF-IDF. However, 21 of these studies, despite employing algorithms such as Support Vector Machines (SVM), Decision Trees, Random Forests, and Logistic Regression, did not specify how negative words were handled during preprocessing. For example, Studies #1 and #3 utilized Logistic Regression with n-grams and TF-IDF for feature extraction but provided no details on managing negations. Similarly, Studies #4 and #18 employed SVM with n-grams and TF-IDF for spam detection without addressing the treatment of negative words, leaving a critical gap in their preprocessing workflows.

Additionally, six of the reviewed studies employed advanced deep learning methods, such as transformer-based architectures like Bidirectional Encoder Representations from Transformers (BERT), which inherently manage contextual meanings and linguistic nuances, including negations, without explicit preprocessing. These models use attention mechanisms and long-range dependencies to capture sentence text, allowing them to interpret the contextual meaning of sentences, such as 'not good,' without needing dedicated preprocessing for negative words. For example, Study #35, utilized BERT, which effectively handled negations through its attention mechanism, capturing the contextual meaning of sentences, without requiring additional preprocessing for negative words. By leveraging these inherent capabilities, the study demonstrated that BERT can accurately process complex linguistic structures like negations, reducing the need for explicit handling.

In fake news detection, accurate sentiment analysis is key to identifying misleading content. Without proper negation handling, models may overestimate the positivity or negativity of content, resulting in misclassifications and undermining the effectiveness of detection systems. The absence of explicit strategies for handling negative words in traditional ML models introduces potential performance biases. Negations, such as "not good", can reverse the sentiment of a sentence, resulting in incorrect interpretations if not properly managed. For instance, in Study #1, the failure to account for negations may have resulted in the misclassification of tweets during crisis events, potentially overlooking critical misinformation that uses negations to alter the sentiment. The advantage of negation preprocessing was observed in our reviewed studies. For example, Study #12 demonstrated that incorporating additional preprocessing steps, such as negation handling and sentiment correction, significantly improved the accuracy of fake news detection when combined with machine learning algorithms like SVM and MultinomialNB.

Moreover, research shows that even advanced DL models can benefit from explicit preprocessing steps, particularly when working with domain-specific language or nuanced features like negations. Kaushik, Hovy, and Lipton (2020) demonstrated that augmenting training data with counterfactual examples, including negations, improved accuracy across a range of NLP tasks. This suggests that even sophisticated models like BERT can be enhanced through preprocessing techniques, improving their ability to interpret complex linguistic structures. *Summary:*

Most reviewed studies did not address negative word handling during preprocessing, highlighting a critical gap and potential performance bias in fake news detection research. Notably, over 72% of all studies, including Study #3, Study #4, and Study #35, did not mention strategies for managing negotiations. In particular, among the 30 studies using traditional machine learning methods,70% failed to integrate strategies for negation handling. Similarly, 83% (5 out of 6) of the studies using advanced deep learning models did not explicitly address negative words, assuming the models' inherent capabilities were sufficient.

Traditional machine learning models are highly reliant on explicit preprocessing to effectively capture linguistic nuances such as negations. Techniques like adding a negation prefix (e.g., transforming "not good" to "NOT_good") or using sentiment lexicons to adjust scores can significantly enhance these models' ability to interpret altered sentiments (Pang & Lee, 2008). Without such strategies, traditional models often misinterpret negated phrases, leading to misclassifications and reduced performance, especially in sentiment-driven tasks like fake news detection. Incorporating sentiment lexicons that adjust scores to account for negations, as well as developing features that specifically detect the presence and impact of negations on sentence meaning, can further improve model accuracy. By adopting these strategies, future research can mitigate potential biases and enhance the effectiveness of machine learning models in tasks such as fake news detection.

While deep learning models, such as transformers, are generally better equipped to manage negations through their attention mechanisms and contextual embeddings (Vaswani et al., 2017), they too have limitations. Incorporating explicit preprocessing steps, such as emphasizing negations, can further boost their performance. Fine-tuning on datasets that contain more examples of negated sentences has been shown to improve the accuracy of these models in interpreting nuanced language structures (Kaushik, Hovy, and Lipton, 2020; Cao et al., 2025; Zhang et al., 2025).

Model Development

*Hyperparameter Tuning (Q3, Q4 & Q5):*

Hyperparameter tuning plays a critical role in optimizing machine learning models and significantly impacts their performance. This section evaluates whether the reviewed studies reported hyperparameters, whether these were optimized, and the consistency of hyperparameter tuning across models. A detailed summary of how these studies report hyperparameters is provided in Table 4. Out of the 36 reviewed studies, 36.1% of them (13 studies) reported hyperparameters and applied tuning techniques such as grid search for all models. For instance, Study #1 reported hyperparameters for models such as SVM, k-nearest neighbors (KNN), decision tree, random forest, and others, applying grid search for optimization across all models to ensure a consistent approach to hyperparameter tuning. Similarly, Study #2, also used grid search to optimize hyperparameters for models including Logistic Regression, SVM, Naive Bayes, KNN, decision tree, random forest, and multi-layer perceptron. This consistency in hyperparameter tuning across models not only ensures that each model is optimized but also allows for fair and accurate comparisons of their performance.

However, 11 studies did not maintain consistency in hyperparameter tuning. For example, Study #7 performed hyperparameter tuning for certain models like SVM and logistic regression but did not apply the same rigor to all models evaluated in their study. Similarly, Study #18 only reported hyperparameters for Random Forest, without tuning for other models. This selective tuning introduces inconsistencies in performance evaluation and model comparison.

In addition, 12 few studies lacked sufficient details on hyperparameter tuning. For example, Studies #6 and #9 neither report hyperparameters nor clarify whether any tuning was performed. This omission hinders the assessment of model optimization and reproducibility.

---

Insert Table 4 about here

---

There are two layers of potential biases in hyperparameter tuning: suboptimal tuning for certain models, and unfair model comparison and selection when not all models have optimized hyperparameters. For example, Study #36 only applied default settings for all models. Additionally, failure to consistently report or optimize hyperparameters can introduce bias when comparing machine learning models, as models that are not equally optimized cannot be fairly compared. For instance, in Studies #3, #7, and #18, only models such as Random Forest and Extreme Gradient Boosting (XGBoost) were fine-tuned by introducing Grid Search with Cross-validation, while other models were left with default settings. This selective tuning can unfairly favor the optimized models, potentially overstating their effectiveness and underrepresenting the true capabilities of the untuned models.

Additionally, a lack of transparency and reproducibility in studies such as Study #6, which did not provide details on hyperparameter tuning, makes it difficult to discern whether a model's poor performance is due to suboptimal tuning or inherent limitations of the model itself. Missing consistency, transparency, and reproducibility can all easily cause biased conclusions about which algorithms are most effective for tasks such as detecting fake accounts or fake news.

The evaluation reveals inconsistency in hyperparameter tuning across the reviewed studies. Approximately 36% of the studies reported hyperparameters and applied tuning techniques consistently across all models, contributing to fair comparisons, and thus, reliable results. In contrast, about 64% of the studies either selectively tuned certain models or failed to report hyperparameter tuning details altogether, introducing potential biases. Models that underwent tuning often appeared to perform better, not necessarily due to their inherent superiority, but because they were optimized. Untuned models may underperform simply due to lack of optimization.

To mitigate potential biases and enhance the reliability of findings, it would be beneficial to consistently apply hyperparameter tuning techniques, such as grid search or random search, across all models evaluated in future research. Providing detailed descriptions of the hyperparameters used and the tuning processes undertaken can improve reproducibility and allow for a more accurate assessment of model performance. Additionally, employing standardized evaluation protocols with consistent evaluation metrics and validation strategies, such as cross-validation, will enable fairer comparisons across models. Assessing the impact of hyperparameter tuning on model performance may also provide valuable insights into the importance of tuning for different algorithms. By adopting these practices, it is possible to reduce biases associated with hyperparameter tuning, leading to more robust and generalizable conclusions about the effectiveness of machine learning models in detecting fake accounts and misinformation on social media platforms.

*Data Partitioning (Q6):*

Proper data partitioning is a critical aspect of machine learning workflows, helping to ensure accurate model evaluation and mitigate overfitting. This section assesses the data partitioning strategies employed in the reviewed studies, particularly whether datasets were appropriately divided into training, validation, and test sets or if cross-validation techniques were used to ensure robust model evaluation.

Among the 36 studies reviewed, data partitioning practices fell into three distinct categories, as outlined in Table 5. A total of 55% of the studies adhered to established machine learning practices by splitting their data into training, validation, and test sets, with performance metrics reported based on separate test sets to evaluate model generalizability (Géron, 2022). For example, Study #1 applied this approach by dividing their dataset into distinct sets for training, validation, and testing, ensuring an unbiased evaluation of their model. Similarly, Study #2 also implemented this split, which contributed to the reliability and validity of their model's performance assessment on new, unobserved data.

Insert Table 5 about here

Moreover, around 28% of the studies employed cross-validation techniques, such as k-fold cross-validation, to enhance the robustness of their evaluations. By employing k-fold cross-validation and similar methods, these studies improved the reliability of their evaluations by testing the model across multiple subsets of the data (Kohavi, 1995). For instance, Study #18 used 10-fold cross-validation to assess their model's performance across multiple data subsets, thereby improving the reliability of their evaluation. Similarly, Study #20 applied 10-fold cross-validation, which helped ensure that their model's performance metrics were not biased by a single train-test split, leading to a more comprehensive evaluation of model effectiveness.

Among the reviewed papers, one study (Study #1) explored the impact of different data splits on model performance. The authors investigated how varying training and testing splits, such as 80-20 and 70-30, influenced the model's predictive ability when combined with five-fold cross-validation. This study provided valuable insights into how different training set sizes can affect the model's generalizability and robustness, especially in the context of misinformation detection during crisis events. By examining these data partitioning strategies, the study offered a detailed understanding of how different training set sizes affect the model's generalizability and robustness, particularly in the context of misinformation detection during crisis events.

However, approximately 17% of the reviewed studies failed to provide sufficient details regarding their data partitioning methods, potentially compromising the reliability and validity of their results. For example, Study #25 did not specify how their dataset was divided into training and test sets or whether cross-validation techniques were employed. The absence of explicit information on data splitting in these studies raises concerns about the validity of their results, as the lack of proper data partitioning can compromise the assessment of a model's ability to generalize to unobserved data (Dieterich, 1995).

Most studies followed best practices in data partitioning, either by clearly dividing data into training, validation, and test sets or by employing cross-validation techniques, ensuring that models were evaluated on unseen data. However, approximately 17% of the studies did not report their data partitioning methods, which poses a significant risk of overfitting. When data partitioning is not clearly defined, models may perform exceptionally well on the training data but poorly on unseen data, leading to inflated performance metrics As Witten and Frank (2002) highlighted, models that are not tested on separate or cross-validated data risk learning specific patterns that do not generalize beyond the training set. This failure to report or utilize adequate partitioning methods results in a lack of transparency and potentially biased conclusions about model performance.

In summary, most reviewed studies adhered to best practices in data partitioning, with 55% employing a clear training/validation/test split and 28% utilizing cross-validation techniques. These approaches enhanced the credibility and applicability of their machine-learning models by providing reliable assessments of model performance and facilitating the development of models that generalize well to unseen data. However, 17% of the studies did not adequately report their data partitioning methods, which may have compromised the validity of their results due to potential overfitting, which occurs when a model learns the training data too closely, including noise and outliers, and fails to perform effectively on new data (Hawkins, 2004; Srivastava et al., 2014). This highlights the importance of proper data partitioning and validation procedures in machine learning research.

Moving forward, it would be beneficial for future research to clearly document data partitioning strategies, including the proportions used for training, validation, and testing. Transparent reporting of these splits is essential for ensuring the reproducibility of results and the fair comparison of models across studies (Xu et al., 2022). Reporting performance metrics based on validation or test sets, rather than solely on training data, can provide a more unbiased assessment of model performance and mitigate overfitting concerns (Cawley & Talbot, 2010; Varma & Simon, 2006). Additionally, adopting advanced partitioning techniques, such as kernel-based subspace clustering methods (Xu et al., 2025), may improve model generalization by capturing complex feature representations. Utilizing and describing cross-validation methods where appropriate can further enhance the robustness of model evaluation by ensuring that results are not biased by a single train-test split. Adopting these practices is likely to improve the transparency, reproducibility, and generalizability of machine learning research, leading to the development of models that are both robust and applicable to real-world scenarios.

Model Evaluation: Evaluation Metrics and Class Imbalance (Q8, Q9 & Q10)

Model evaluation is critical in the context of detecting fake news and fake accounts, particularly due to the significant challenge of class imbalance in the datasets. Typically, real content greatly outnumbers fake content, leading to skewed distributions where standard models may perform well on majority classes (i.e., real news) but fail to correctly classify minority classes (i.e., fake news). This imbalance necessitates special methods to ensure that models are not biased towards the majority class. Two key techniques for addressing class imbalance in machine learning include reweighting and resampling. Reweighting assigns higher penalties for misclassifying instances from the minority class, thereby forcing the model to focus more on these cases during training. Resampling, on the other hand, involves adjusting the dataset by either oversampling the minority class or undersampling the majority class to create a more balanced distribution. In addition to these techniques, robust evaluation metrics are crucial when dealing with imbalanced data. Metrics such as precision, recall, F1 score, or Area Under the Receiver Operating Characteristic Curve (AUROC) are more informative in such cases than simple accuracy, especially in imbalanced scenarios.

In the context of fake identification, precision measures how many of the news items flagged as fake are fake, helping reduce false positives and maintain credibility. Recall, on the other hand, focuses on how well the model identifies actual fake news, ensuring that misinformation is detected and prevented from spreading. However, balancing recall with precision is essential, as high recall alone can lead to many false positives (Bishop, 2006). The F1 score combines precision and recall into a single metric, providing a balanced view of the model's performance. Additionally, the AUROC evaluates the model's ability to distinguish between fake and real news across different thresholds, making it especially useful for handling imbalanced datasets and assessing the overall effectiveness of fake news detection.

Across the reviewed studies, many recognized the importance of these advanced metrics, with 86.1% of the studies (Studies #1, #3, #5, #7, #8, #10, and #12—#17, #19—#36) incorporating F1 score, precision, recall, or AUROC in their evaluations, moving beyond accuracy, which can be misleading in imbalanced datasets. For example, Study #1 utilized precision, recall, and the macro-averaged F1 score as evaluation metrics. By doing so, they acknowledged the inherent class imbalance in their dataset and provided a more accurate reflection of the model's ability to handle both majority and minority classes effectively.

Several studies (Studies #3, #4, #6, #7, #9, #11, #35) addressed this issue by balancing their datasets, either naturally or through specific data preprocessing techniques. For instance, Study #35 used a dataset with near-balanced classes, where real news constituted approximately 52% and fake news about 48%, mitigating the need for class imbalance-specific preprocessing steps. In cases of more pronounced imbalance, such as in Study #3, the researchers balanced their dataset by randomly selecting non-spammer users to match the number of fake accounts (i.e., undersampling the majority class). This approach improved the model's ability to detect both fake and real accounts. In contrast, Study #7 employed oversampling of the minority class using the SMOTE-NC (Synthetic Minority Over-sampling Technique for Nominal and Continuous data) algorithm. This approach helped mitigate the bias introduced by imbalanced datasets and enhanced the model's ability to detect both fake and real accounts. In addition to resampling techniques, some studies also employed reweighting strategies. For example, Study #4 dealt with class imbalance in a phishing detection system by assigning more weights to the minority phishing class. The dataset was heavily skewed towards legitimate tweets, but by applying these weights, the researchers were able to balance the prediction error and minimize the overall error rate. This reweighting approach ensured that phishing tweets, which were much rarer, were still given adequate attention by the model.

Furthermore, some studies (e.g., Study #7) went beyond data preprocessing techniques by incorporating preferred evaluation metrics such as precision, recall, and F1 scores to provide a more balanced and informative assessment of model performance.

However, not all studies (Studies #2, #9, and #18) adequately addressed the issue of class imbalance. For example, Study #9 relied primarily on accuracy as evaluation metrics, without implementing any resampling techniques or strategies to address class imbalance. This oversight suggests that researchers did not fully recognize the potential bias introduced by imbalanced data, leading to inflated performance metrics when using accuracy alone. Similarly, Study #18 depend heavily on accuracy as the main evaluation metric without acknowledging or addressing the skewed class distribution in their dataset. This failure to consider class imbalance undermines the validity of their results, as accuracy can be artificially inflated by the overrepresentation of majority-class examples. In contrast, Study 2 did recognize the issue of class imbalance but attempted to address it only through basic normalization techniques like Min-Max scaling, which does not solve the imbalance problem. While the researchers were aware of the issue, their approach may not have been sufficient, as more effective methods like resampling or reweighting would have been necessary to properly handle the imbalanced data, potentially limiting the reliability of their model's performance evaluation.

Overall, 86.1% of the reviewed studies did not adequately address the class imbalance inherent in fake news datasets. While some researchers effectively employed preprocessing techniques and utilized preferred evaluation metrics, others continued to rely on outdated practices, such as focusing on accuracy without considering the effects of class imbalance. To enhance the robustness and reliability of fake news detection models, future research should adopt more comprehensive approaches, including data preprocessing methods such as resampling, reweighting, and the consistent use of preferred metrics like recall, precision, F1 score, and AUROC. Incorporating these strategies will lead to a more accurate and nuanced evaluation of model performance, especially in scenarios with imbalanced scenarios.

Reporting: Transparency and Completeness:

In evaluating the transparency and completeness of reporting in the reviewed studies, we assessed whether they provided sufficient details about their data collection, preprocessing, model development, hyperparameter tuning, and evaluation processes. Transparent reporting is essential for reproducibility and allows for critical evaluation of methodologies and findings.

Out of the 36 studies reviewed, 20 studies (55.6%) demonstrated a high level of transparency. These studies provided comprehensive details about their methodologies, including data sources, sampling methods, preprocessing steps, model architectures, hyperparameter tuning processes, and evaluation metrics. For example, Study #1 meticulously documented their sampling methods, dataset size, preprocessing techniques, model parameters, and evaluation metrics, facilitating reproducibility and critical assessment. Study #4 provided detailed descriptions of their dataset collection, feature extraction process, hyperparameter tuning using grid search for all models, and provided code snippets or references to repositories.

Conversely, 16 studies (44.4%) lacked transparency in critical areas: 12 studies (33.3%) failed to fully describe data sources, sampling criteria, or preprocessing steps, limiting assessment of data quality and representativeness (e.g., Study #18 lacked details on non-spam user sampling). In 23 studies (63.9%), hyperparameter tuning was inconsistently applied or undocumented, with some models optimized (e.g., SVM in Study #7) while others were not. Additionally, 9 studies (25%) did not clearly report evaluation metrics or justify their choices, particularly in handling class imbalance, as seen in Study #9, which relied on accuracy without addressing imbalance issues.

The lack of comprehensive reporting in many studies introduces several biases: Performance bias arises from inadequate documentation of hyperparameter tuning and model development, potentially leading to overestimation of a model's effectiveness due to overfitting. Reproducibility issues emerge when insufficient methodological details prevent other researchers from replicating or building on the work, hindering scientific progress. Interpretation bias occurs when there is insufficient reporting on data preprocessing and handling of class imbalance, making it difficult to assess the validity of the results and potentially leading to misleading conclusions.

Approximately 55.6% of the reviewed studies provided transparent and comprehensive reporting, enhancing the credibility and reproducibility of their findings. However, 44.4% of the studies lacked sufficient detail in critical areas such as data collection, preprocessing, hyperparameter tuning, and model evaluation. This lack of transparency introduces potential biases and hampers the ability to critically assess and replicate the studies.

To mitigate these issues, future research should prioritize comprehensive methodological reporting by providing detailed descriptions of data sources, sampling methods, preprocessing steps (including handling of linguistic nuances like negations), model architectures, and hyperparameter tuning processes. Additionally, it is crucial to ensure transparent evaluation procedures by justifying the choice of evaluation metrics, particularly in imbalanced datasets, and reporting all relevant metrics (precision, recall, F1 score, AUROC). Describe validation methods, including data partitioning strategies. Clear descriptions of validation methods and data partitioning strategies are equally important. Furthermore, studies should include an acknowledgment of potential biases and limitations, such as non-representative sampling or class imbalance issues, and explain the strategies used to mitigate them.

Finally, enhancing reproducibility should be a key goal, with researchers encouraged to share datasets and code repositories whenever possible. Following established reporting guidelines, such as the Transparent Reporting of a Multivariable Prediction Model for Individual Prognosis or Diagnosis (TRIPOD) statement, can further strengthen the reliability and transparency of research in machine learning applications for fake news detection. By adopting these practices, researchers can improve the reliability and impact of their studies, contributing to the advancement of machine learning applications in fake news detection.

Result Summary

This section provides a systematic review of studies using ML and DL models for fake news detection on social media. The analysis highlights key insights into model effectiveness while identifying significant limitations and biases impacting performance and generalizability. Biases were found across the machine learning lifecycle, including non-representative sampling, inadequate handling of linguistic nuances, improper hyperparameter tuning, and over-reliance on accuracy for imbalanced datasets, which reduces model effectiveness in real-world scenarios.

To address these issues, future research should focus on improving representative sampling by incorporating diverse social media platforms, languages, and demographics. Standardizing preprocessing techniques, especially for handling linguistic features and class imbalance, is essential for model improvement. Consistent hyperparameter tuning and transparent reporting will further enhance model optimization and comparability.

Addressing class imbalance with appropriate evaluation metrics like F1 score and recall, rather than accuracy, is critical for robust model performance. Increased transparency and reproducibility, including sharing datasets and code, will support validation and advancement in the field. Lastly, exploring advanced models and integrating multimodal data (text, images, videos) can enhance detection capabilities.

In conclusion, overcoming these challenges will lead to more accurate, generalizable machine learning models for fake news detection, strengthening efforts to combat misinformation on social media and fostering trustworthy online environments.

### Discussion

The proliferation of misinformation and deceptive content on social media platforms has necessitated the development of automated detection methods. ML and DL models have shown considerable promise in identifying fake news, spam, and fake profiles by analyzing vast amounts of data and capturing complex patterns. These models, particularly DL architectures, can handle nuanced language and contextual cues, making them powerful tools in the fight against misinformation. However, this systematic review of 36 studies employing ML and DL techniques reveals several key limitations and biases, such as selection bias, inconsistent data preprocessing, class imbalance, and poor hyperparameter tuning, which hinder the full potential of these models in real-world applications.

Key Findings and Identified Biases

The reviewed studies utilized a diverse array of ML and DL models to detect deceptive content. Traditional supervised learning algorithms such as Random Forest (used in 17 studies), Support Vector Machines (SVM) (16 studies), Naive Bayes (12 studies), and Logistic Regression (14 studies) were prominently featured due to their effectiveness in classification tasks. Deep learning models, particularly Artificial Neural Networks (ANN) and Long Short-Term Memory (LSTM) networks, were also employed in several studies (8 and 2 studies, respectively). These models demonstrated enhanced capabilities in capturing complex patterns and contextual information within unstructured data (Wang et al., 2019). For example, studies using LSTM networks reported F1 scores exceeding 90%, showcasing their effectiveness in handling sequential textual data prevalent in social media posts. Overall, these machine learning and deep learning models have proven useful in detecting fake news, fake accounts, and other types of misinformation, showing promising results in enhancing accuracy and robustness in a range of detection tasks across different platforms.

The majority of studies focused on Twitter (19 studies) and Instagram (9 studies), reflecting the prominence of these platforms in social discourse and the availability of data. Facebook and Reddit were less represented, with 3 studies each, and a few studies explored multiple platforms or other social media like Weibo. This platform focus introduces selection bias, limiting the generalizability of findings across diverse social media platforms, which exhibit different user behaviors and content types. The types of deceptive content analyzed varied, with a significant focus on fake profiles and fake accounts (15 studies) and general fake news detection (12 studies). Spam and scam detection were addressed in 5 studies, while phishing detection and health-related misinformation were less commonly explored. Only one study specifically targeted misinformation during crisis events, suggesting a gap in research for this critical area of public concern.

Biases were identified throughout the entire lifecycle of machine learning applications, from data collection to model evaluation. In data sampling, approximately 55% of the studies exhibited selection bias by focusing on specific events or user groups, limiting the generalizability of their findings. Class imbalance was another major issue, with many studies failing to properly address the disproportion between real and deceptive content, resulting in models biased toward the majority class. In data preprocessing, 86% of the studies did not adequately handle linguistic nuances, such as negations, which can lead to misinterpretation of text and reduced model accuracy. During model development, inconsistent hyperparameter tuning was found in 64% of the studies, leading to suboptimal model performance and making fair comparisons between models difficult. Finally, model evaluation practices showed bias, as many studies relied heavily on accuracy, a metric not suited for imbalanced datasets, rather than using more appropriate measures like precision, recall, and F1 score. These issues collectively impact the reliability and applicability of the models in detecting deceptive content in real-world scenarios.

Limitations

This systematic review provides valuable insights into the application of machine learning and deep learning models for fake news detection on social media platforms, but several limitations should be acknowledged. First, the review focuses on a specific set of studies that utilized machine learning for detecting fake news, spam, and fake profiles. This limited scope means it may not capture all relevant studies, particularly those published in less accessible databases or in languages other than English. Moreover, the rapid advancement of machine learning techniques, especially in natural language processing and deep learning, presents another limitation. Some of the methodologies discussed may already be outdated, with newer approaches potentially offering improved performance and addressing challenges like bias and class imbalance more effectively.

Another limitation lies in the focus of most reviewed studies on textual data. While text analysis is central to fake news detection, many studies did not explore multimodal approaches that integrate other forms of content such as images, videos, and metadata. Given the growing prevalence of multimedia content in online misinformation, this narrow focus may understate the potential of more comprehensive, multimodal detection techniques. Additionally, the variability in evaluation standards across the studies, including differences in preprocessing techniques, hyperparameter tuning, and performance metrics, makes it difficult to directly compare their results. This inconsistency hampers the ability to draw broad, definitive conclusions about the most effective models and approaches.

While the review focuses primarily on technical issues like bias and class imbalance, it does not extensively address the ethical considerations involved in deploying machine learning models for fake news detection. Important concerns such as privacy, fairness, and the potential for unintended societal consequences were outside the scope of this review but are critical factors that should be considered when developing automated misinformation detection systems. Finally, the reviewed studies mostly focused on retrospective analyses of pre-collected data, with limited exploration of the challenges involved in real-time detection. Since real-time detection is vital in curbing the immediate spread of misinformation, this gap highlights an area where future research should concentrate.

By acknowledging these limitations, the review underscores the need for more comprehensive approaches in future research, such as the development of multimodal detection techniques, the adoption of standardized evaluation methods, and the exploration of real-time detection capabilities. For example, recent work on AI-driven alt text generation demonstrates the advantages of integrating textual and visual information to enhance content analysis and accessibility (Shen et al., 2024). Such multimodal integrations could provide richer context for misinformation detection. Additionally, integrating ethical considerations into research practices will be essential for ensuring the responsible deployment of machine learning models to combat misinformation effectively.

Concluding Remark

This systematic review highlights the potential of machine learning and deep learning models in detecting fake news and other forms of deceptive content on social media platforms. However, the review also underscores several critical limitations in current approaches, particularly in terms of selection bias, class imbalance, inadequate handling of linguistic nuances, inconsistent hyperparameter tuning, and reporting practices. These challenges undermine the generalizability and robustness of the models reviewed, signaling a need for improved methodologies.

The limitations identified in this review highlight the need for future research to focus on more representative sampling, improved handling of data imbalances, and the adoption of multimodal approaches. Multimodal learning, which has been successfully applied to various classification tasks, such as protest activity recognition (Lu et al., 2023), offers valuable contextual information that can enhance misinformation detection. Furthermore, cross-boundary AI innovation has demonstrated that integrating heterogeneous data sources—including text, images, and metadata—can significantly improve model robustness and generalization in complex real-world applications (Gao et al., 2024). As real-time detection and ethical considerations become increasingly critical in combating misinformation, interdisciplinary research and practical implementations will be essential to ensuring both the effectiveness and fairness of automated detection systems.

By addressing these limitations, future research can advance the field of fake news detection, making machine learning models more reliable, transparent, and applicable across a broader range of real-world scenarios. This progress is crucial to curbing the spread of misinformation and ensuring a more trustworthy online environment.

**Appendix 1. Reviewed Studies on Machine Learning Models for Fake News/Profile Detection on Social Media**

Insert Table A1 about here

Table 1. Key Data Extraction Categories for Systematic Review

| Category | Details |
| --- | --- |
| Study Details | Title, Authors, Year of Publication, Journal or Source, DOI or URL |
| Research Objectives | Purpose of the Study, Research Questions or Hypotheses |
| Methodological Aspects | Study Design, Settings, Sample Sizes, Inclusion and Exclusion Criteria, Data Collection Methods, ML/DL Models Employed |
| Criteria Applied | Data included, e.g., publicly available tweets, specific language posts |
| | Data excluded, e.g., private or insufficiently detailed posts |
| Performance Metrics | Metrics Used (e.g., Accuracy, Precision, Recall, F1-score, AUROC, etc.) |
| Bias Evaluation | Data Collection and Preprocessing, Model Development and Tuning, Model Evaluation and Reporting. |
| Additional Information | Confounding Factors, Study Limitations, Ethical Considerations, Funding Sources |

Table 2. Bias Evaluation Questions for Each Domain

| Domain | Evaluation Questions |
| --- | --- |
| Sample Selection and Representativeness | Q1. What is the sample used in this study, including the platform, sampling criteria, and sampling method? |
| | Q2. Does the sample represent the target population of social media users or posts? |
| Data Preprocessing | Q3. Did the study specify its approach to handling negative words when using traditional or machine learning methods for sentiment analysis? |
| Model Development | Q4. Did this study report hyperparameters? |
| | Q5. If reported, did this study tune (optimize) hyperparameters or use default settings? |
| | Q6. If tuned hyperparameters in this study, was this done on all models mentioned in the study? |
| Model Evaluation | Q7. Did the study divide the dataset into training, validation, and test sets, and were the reported metrics based only on training data? |
| | Q8. What evaluation metric was used in this study? |
| | Q9. Is the evaluation metric appropriate for this context (i.e., class-imbalanced settings)? |
| | Q10. If the study used accuracy as an evaluation metric, did it mention preprocessing steps to address class imbalance? |

Table 3. Sampling Approaches Across Reviewed Studies

| Sampling Approach | Number (%) of Studies | Studies # |
|---|---|---|
| Specific Events or User Groups | 20 (55%) | #1, #2, #3, #5, #7, #8, #9, #10, #12, #13, #14, #16, #17, #18, #20, #21, #27, #31, #33, #34, #36 |
| General Sampling with Single Platform | 10 (28%) | #4, #6, #15, #19, #24, #26, #28, #29, #30, #32 |
| Comprehensive Multi-Platform Sampling | 6 (17%) | #11, #22, #23, #25, #35 |

Table 4. Hyperparameter Reporting and Tuning Practices in Reviewed Studies

| Hyperparameter Reporting | Number (%) of Studies | Studies # |
|---|---|---|
| Reported & Tuned for All Models | 13 (36.1%) | #1, #2, #10, #11, #12, #13, #17, #21, #22, #23, #24, #33, #35 |
| Reported but Partially Tuned | 11 (30.6%) | #3, #4, #5, #7, #8, #18, #19, #29, #30, #34, #36 |
| Not Reported or Tuned | 12 (33.3%) | #6, #9, #14, #15, #16, #20, #25, #26, #27, #28, #31, #32 |

Table 5. Summary of Data Partitioning Practices Across Reviewed Studies

| Data Partitioning Practices | Number (%) of Studies | Studies # |
|---|---|---|
| Training/Validation/Test | 20 (55%) | #1, #2, #3, #4, #5, #6, #7, #8, #9, #10, |

| Split | | #11, #12, #13, #14, #15, #16, #17, #29, #31, #35 |
|---|---|---|
| Cross-validation without Traditional Split | 10 (28%) | #18, #19, #20, #21, #22, #23, #24, #30, #34, #36 |
| Inadequate or Unreported Partitioning | 6 (17%) | #25, #26, #27, #28, #32, #33, |

Table A.1. Titles and Authors of Reviewed Studies on Machine Learning Models in Detecting Deceptive Activities on Social Media

| Index | Title of paper | Reference |
|---|---|---|
| 1 | Monitoring Misinformation on Twitter During Crisis Events: A Machine Learning Approach | Hunt et al. (2022) |
| 2 | Classification of Fake, Bot, and Real Accounts on Instagram Using Machine Learning | Tunç et al. (2022) |
| 3 | Scam Detection in Twitter | Chen et al. (2014) |
| 4 | PhishAri: Automatic Realtime Phishing Detection on Twitter | Aggarwal et al. (2012) |
| 5 | A Survey on Machine Learning Algorithms for Detecting Fake Instagram Accounts | Anklesaria et al. (2021) |
| 6 | Application of Machine Learning Techniques in Detecting Fake Profiles on Social Media | Bhattacharya et al. (2021) |
| 7 | Instagram Fake and Automated Account Detection | Akyon & Kalfaoglu (2019) |
| 8 | Supervised Machine Learning Algorithms to Detect Instagram Fake Accounts | Ekosputra et al. (2021) |
| 9 | Fake Instagram Profile Identification and Classification Using Machine Learning | Harris et al. (2021) |
| 10 | Fake News Detection in Social Networks Using Machine Learning and Deep Learning: Performance Evaluation | Han & Mehta (2019) |
| 11 | Enhancing the Detection of Fake News in Social Media Based on Machine Learning Models | Amin et al. (2023) |
| 12 | Fake News Detection on Reddit Utilizing CountVectorizer and Term Frequency-Inverse Document Frequency with Logistic Regression, MultinomialNB, and Support Vector Machine | Patel & Meehan (2021) |
| 13 | Detecting Fake News on Twitter Using Machine Learning Models | Cueva et al. (2020) |
| 14 | Machine Learning Algorithm-Based Model for Classification of Fake News on Twitter | Nikam & Dalvi (2020) |
| 15 | Detection of Fake Profiles on Twitter Using Hybrid SVM Algorithm | Kodati et al. (2021) |

| 16 | The Detection of Fake Messages Using Machine Learning | Looijenga (2018) |
|---|---|---|
| 17 | Detecting Fake News Using Machine Learning and Deep Learning Algorithms | Abdullah All et al. (2019) |
| 18 | A Machine Learning Approach for Twitter Spammers Detection | Meda et al. (2014) |
| 19 | Scalable Learning Framework for Detecting New Types of Twitter Spam with Misuse and Anomaly Detection | Choi et al. (2024) |
| 20 | Classification of Instagram Fake Users Using Supervised Machine Learning Algorithms | Purba et al. (2020) |
| 21 | Approaches in Fake News Detection: An Evaluation of Natural Language Processing and Machine Learning Techniques on the Reddit Social Network | Shariff et al. (2022) |
| 22 | Multiple Features Based Approach for Automatic Fake News Detection on Social Networks Using Deep Learning | Sahoo & Gupta (2021) |
| 23 | Social Networks Fake Account and Fake News Identification with Reliable Deep Learning | Kanagavalli & Priya (2022) |
| 24 | Weakly Supervised Learning for Fake News Detection on Twitter | Helmstetter and Paulheim (2018) |
| 25 | Survey on Fake Profile Detection on Social Sites by Using Machine Learning Algorithm | Patel et al. (2020) |
| 26 | Twitter Fake Account Detection | Erşahin et al. (2017) |
| 27 | Using Linguistics and Psycholinguistics Features in Machine Learning for Fake News Classification Through Twitter | Lee & Chua (2022) |
| 28 | Detecting Fake News Using Machine Learning Algorithms | Bharath, G. et al. (2021) |
| 29 | Hybrid approach for detection of malicious profiles in twitter | Sahoo and Gupta (2019) |
| 30 | An effective security alert mechanism for real-time phishing tweet detection on Twitter | Liew (2019) |
| 31 | Credibility detection on twitter news using machine learning approach | Azer, et al. (2021) |
| 32 | Machine learning implementation for identifying fake accounts in social network | Raturi (2018) |
| 33 | Automatic detection of fake profile using machine learning on instagram | Meshram (2021) |
| 34 | Validating Machine Learning Algorithms for Twitter Data Against Established Measures of Suicidality | Braithwaite et al. (2016) |
| 35 | Towards COVID-19 fake news detection using transformer-based models | Alghamdi et al., (2023) |
| 36 | A machine learning approach predicts future risk to suicidal ideation from social media data | Roy et al. (2020) |

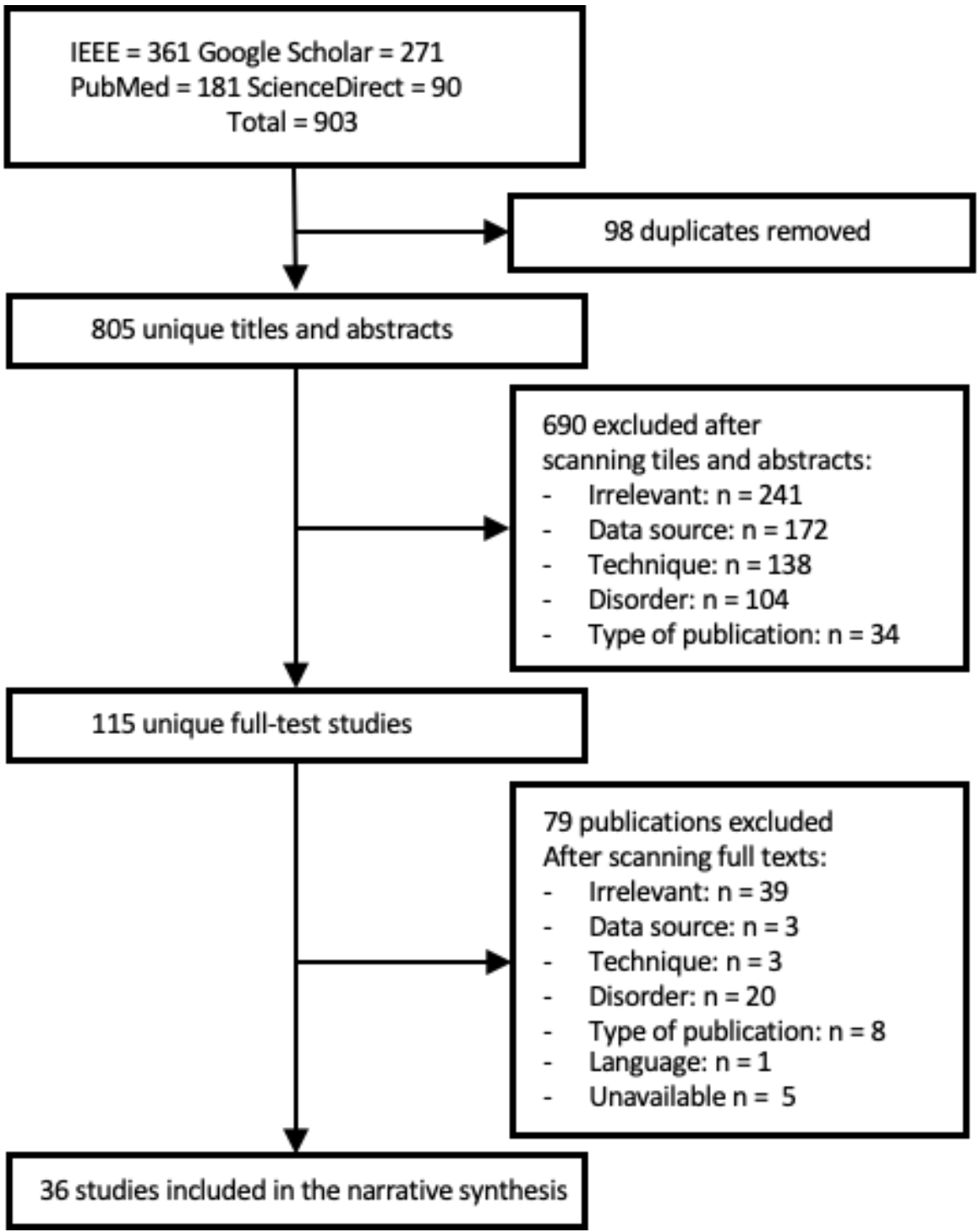

Figure 1: PRISMA Flow Diagram of Study Selection Process for Systematic Review on Machine Learning Models in Detecting Deceptive Activities on Social Media